\documentclass{article}

\usepackage{arxiv}

\usepackage[utf8]{inputenc} 
\usepackage[T1]{fontenc}    
\usepackage{hyperref}       
\usepackage{url}            
\usepackage{booktabs}       
\usepackage{amsfonts}       
\usepackage{nicefrac}       
\usepackage{microtype}      
\usepackage{lipsum}
\usepackage{balance}
\usepackage{graphicx}
\usepackage{placeins}
\usepackage{float}
\usepackage{subfigure}
\usepackage{verbatim} 
\usepackage{amsmath,amssymb,amsfonts}
\usepackage{algorithmic}
\usepackage{algorithm}
\usepackage{multirow}
\usepackage{listings}

\title{Automatic Construction of Multi-layer Perceptron Network from Streaming Examples \thanks{This paper has been accepted for publication in CIKM 2019. The source code is available in \url{https://www.researchgate.net/publication/335813786_NADINE_Code_M_File}.}}

\author{
	Mahardhika Pratama \\
	School of Computer Science and Engineering\\
	Nanyang Technological University, Singapore \\
	\texttt{mpratama@ntu.edu.sg} \\
	\And
	Choiru Za'in \\
	Department of Computer Science and Information Technology\\
	La Trobe University, Australia \\
	\texttt{c.zain@latrobe.edu.au} \\
	\And
	Andri Ashfahani \\
	School of Computer Science and Engineering\\
	Nanyang Technological University, Singapore \\
	\texttt{andriash001@ntu.edu.sg} \\
	\And
	Yew Soon Ong \\
	School of Computer Science and Engineering\\
	Nanyang Technological University\\
	50 Nanyang Avenue, Singapore \\
	\texttt{asyong@ntu.edu.sg} \\
	\And
	Weiping Ding\\
	School of Computer Science and Engineering\\
	Nantong University, China\\
	\texttt{dw9988@163.com} \\
}

\begin{document}
	\maketitle
	
	\begin{abstract}
		Autonomous construction of deep neural network (DNNs) is desired for data streams because it potentially offers two advantages: \textbf{proper model's capacity} and \textbf{quick reaction to drift and shift}. While self-organizing mechanism of DNNs remains an open issue, this task is even more challenging to be developed for standard multi-layer DNNs than that using the different-depth structures, because addition of a new layer results in information loss of previously trained knowledge. A Neural Network with Dynamically Evolved Capacity (NADINE) is proposed in this paper. NADINE features a fully open structure where its network structure, depth and width, can be automatically evolved from scratch in the online manner and without the use of problem-specific thresholds. NADINE is structured under a standard MLP architecture and the catastrophic forgetting issue during the hidden layer addition phase is resolved using the proposal of soft-forgetting and adaptive memory methods. The advantage of NADINE, namely elastic structure and online learning trait, is numerically validated using nine data stream classification and regression problems where it demonstrates performance's improvement over prominent algorithms in all problems. In addition, it is capable of dealing with data stream regression and classification problems equally well.
	\end{abstract}

	
	\section{Introduction}
	\label{submission}
	The major topic in the incremental learning (IL) of data stream lies in the adaptation of the new concept without suffering from the catastrophic forgetting problem \cite{kirkpatrick2016overcoming}. This property is absent in the vast majority of conventional DNNs having a fixed structure and an offline working principle. It is often assumed that a full access of overall system conditions is observable and concept drifts \cite{Gamaconceptdrift} impose a retraining phase from scratch which catastrophically replaces old knowledge with a new one \cite{Learn++NSE}. Moreover, a static network structure is hindered by the fact where it is difficult to estimate the right network complexity before process runs. Data streams must be also handled in the sample-wise manner ideally in the single scan to assure scalability. In other words, the iterative training process over a number of epochs cannot keep pace with rapidly changing data stream applications.

	Incremental Learning of data streams is defined as a learning problem of sequence of tasks where a new task has to be embraced in such a way to minimize performance degradation due to changing learning environments. The elastic consolidation weight (ECW) method is proposed in \cite{kirkpatrick2016overcoming} where it overcomes the catastrophic forgetting issue by preventing a large deviation of output weights from the old one. Online deep learning (ODL) is proposed in \cite{OnlineDeepLearning} which introduces the idea of hedging. The hedging concept opens a direct connection of hidden layer to output layer with a connective weight and shares some similarity to the concept of weighted voting scheme in the ensemble learning literature. Nevertheless, these approaches are crafted under a fixed network structure. 
	
	The concept of dynamic structure has started to emerge recently. In \cite{Zhou_incrementallearning}, the incremental learning scenario of denoising auto-encoder (DAE) is proposed using the feature similarity concept.  Progressive neural network (PNN) is designed in \cite{progressiveneuralnetworks} where it freezes the network parameters of previous tasks while introducing new column to embrace the current task. Because the network complexity of PNN proportionally grows as the number of tasks, the idea of dynamically expandable network is put forward in \cite{DeepExpandable} and works with the splitting/duplicating mechanism of hidden units. The three approaches, however, rely on a predefined threshold and still operate with a fixed depth. Note that a very deep network structure compromises the network complexity aspect and is hard to train due to the vanishing gradient problem \cite{verydeep,IMM2012}. Autonomous deep learning (ADL) \cite{ADL} can be seen as the recent advance of self-organizing deep neural networks addressing the aforesaid problems. Nevertheless, ADL is crafted under the different-depth structure where every layer is assigned with a softmax layer and the final output is produced from a weighted voting mechanism. Such structure limits the application of ADL only for classification problem while the use of local softmax layer incurs extra network parameters.    
	
	A novel deep neural network (DNN), namely Neural Networks with Dynamically Evolved Capacity (NADINE), is proposed for handling evolving data streams. It characterizes a re-configurable structure where hidden units are dynamically grown and pruned while the depth of network structure grows or shrinks on demand. Note that the most challenging case is considered here where the training process starts with the absence of an initial structure. Moreover, NADINE is framed under the standard MLP network without any bypass connection where its structural evolution notably insertion of a new hidden layer incurs the catastrophic forgetting problem. The advantage of such structure is capable of performing the non-classification tasks. NADINE addresses the issue of catastrophic forgetting during the structural learning phase with the notion of adaptive memory and the soft forgetting method which warrants right balance between old and new knowledge. The salient property of NADINE is elaborated as follows:
	
	\textbf{Adjustment of Network Width}: NADINE puts forward Network Significance (NS) method which informs statistical approximation of network contribution. It is formulated as a limit integral approximation of generalization error under normal distribution leading to the prominent bias and variance decomposition. A new hidden unit is introduced in the case of underfitting, high bias, whereas an inconsequential hidden unit is pruned in the case of overfitting, high variance. All of which are carried out in the one-pass manner and are extendable to different DNN variants. This strategy goes one step ahead of existing lifelong DNNs adapting the network structure in respect to network loss and being over-dependent on the user-defined threshold. It must be also understood that DNN cannot be approached with the clustering methods as popular in the RBF networks \cite{MRAN}. 
	
	\textbf{Expansion of Network Depth}: The depth of NADINE network structure is adjustable using the concept drift detection mechanism identifying the presence of real drift in the data stream - changes in both feature and target spaces \cite{Gamaconceptdrift}. A new hidden layer is inserted given that a drift is signalled. This strategy aims to set hidden layers as a representation of different concepts in data stream. 
	The drift detection mechanism of NADINE advances the Hoeffding bound method \cite{Pesaranghader2018} using the concept of an adaptive sliding window \cite{drift}. It also goes one step ahead from \cite{drift} where the real drift, change of both input and target space, is considered rather than merely the covariate drift, change of input space. Note that addition of a new layer requires a particular mechanism to overcome information loss due to introduction of an untrained layer as a last layer governing the final network output.    
	
	\textbf{Solution of Catastrophic Forgetting}: The catastrophic forgetting problem is a challenging problem here due to the nature of MLP structure. Such structure is, however, general in nature and is applicable to wide range of applications: regression, control, reinforcement learning, etc. It is addressed via two mechanisms: adaptive memory and soft forgetting mechanism. The concept of adaptive memory utilizes a selective sampling method extracting salient samples. The salient samples reveals the underlying structure of data distribution and are stacked in the adaptive memory. These data samples are combined with new samples and are prepared to create a new hidden layer if a drift is identified. It assures relevance of network structure to current data trend without any compromise of old system dynamics. Another strategy for catastrophic forgetting is crafted with the concept of soft forgetting. That is, the learning rates of the SGD method are set to be fully adaptive and are unique per layer. The setting of learning rate is based on the relevance of a hidden layer to current concept where an irrelevant layer is frozen by setting its learning rate to a small value and vice versa.
	
	Novel contribution of NADINE is summarized as follows: 1) NADINE offers automatic construction method of DNNs from online data streams under the classic MLP architecture. To the best of our knowledge, this work is the first attempt in evolving both the network width and depth of MLP network with a solution to the issue of catastrophic forgetting during the online structural evolution; 2) The adaptive memory and soft forgetting methods are proposed to specifically cope with information loss during addition of a new hidden layer; 3)  NADINE is capable of not only dealing with the classification problems, but also regression problems. It is also shown that NADINE is capable of evolving its network structure on demand and in a non-ad-hoc manner while fully working in one-pass learning mode. Numerical study in nine prominent data stream problems with non-stationary characteristics demonstrates that NADINE performs favourably compared to recently published algorithms and handles the regression problems as well as the classification problems.
	\section{Problem Formulation}
	Incremental learning (IL) of data streams is defined as a learning strategy of sequentially arriving tasks $B_K=[B_1,B_2,B_k,...,B_K]$ where $K$ denotes the number of tasks \cite{GamaDataStream}. DNN is supposed to handle $B_k$ which may be drawn from different data distributions - the concept drift. In other words, there may exist a change of the joint class posterior probability $P(Y_t,X_t) \neq P(Y_{t-1},X_{t-1})$. Moreover, the type, rate and severity of concept changes vary. In practise, data streams arrive with the absence of the true class label $B_k=X_k\in\Re^{T\times n}$ where $T,n$ respectively stand for the number of data points and the input dimension. Operator labelling effort is usually solicited to associate an input data sample $X_t$ to its true class label $Y_k\in\Re^{T}$. The one-hot encoding scheme is applied leading to $Y_k\in\Re^{T \times m}$ where $m$ is the output dimension. $Y_k$ is a real-valued variable in the context of regression problem. To this end, our simulation is run under \textbf{the prequential test-then-train procedure} - a standard evaluation procedure of data stream methods \cite{GamaDataStream}. That is, the testing process is carried out first prior to the training process and this process takes place over the number of tasks $K$. Another challenge of incremental learning lies in achieving a tradeoff between plasticity and stability without being trapped to catastrophic forgetting situation \cite{kirkpatrick2016overcoming}. 
	
	NADINE is formed as a standard deep feed-forward neural network with the sigmoid activation function $s(X_t W_d+b_d)$ and the softmax layer as the last layer which assigns output probability to each class $softmax(x_o)=\frac{exp(x_o)}{\sum_{o=1}^{m}exp(x_o)}$. No direct connection or highway is inserted in hidden layer making the catastophic forgetting difficult to be addressed. This architecture, however, can be generalized conveniently to non-classification tasks. On the other hand, no softmax layer is attached in the last layer in the regression problem where the output weight directly links the last hidden layer to the target space. $W_d\in\Re^{n \times R_d}$ and $b_d\in\Re^{R_d}$ denote the connective weight and bias of $d-th$ hidden layer where $R_d$ is the number of hidden nodes of $d-th$ hidden layer. NADINE features a dynamically expandable structure where its network structure is self-constructed from data streams. It is capable of initiating its training process from scratch with the absence of initial structure $R_d=1,d=1$. The hidden unit evolves on demand using the NS method - approximation of network statistical contribution. A new hidden unit is appended in the case of underfitting $R_d=R_d+1$ whereas the overfitting situation triggers the hidden unit pruning mechanism $R_d=R_d-1$. 
	
	The drift detection module is designed under the concept of Hoeffding's bound method making use of the adaptive windowing scheme \cite{drift}. A new hidden layer is inserted $D=D+1$ if a drift is signalled. Nevertheless, a creation of hidden layer should be carefully undertaken to address the catastrophic forgetting issue \cite{kirkpatrick2016overcoming}. The concept of adaptive memory is applied in the hidden layer growing process and is inspired by the experience-replay method \cite{Schaul2016}. It extracts key samples disclosing hidden structure of data streams. The salient samples are replayed along with new data samples if a new hidden layer is appended. Note that this strategy aligns with the recent finding of \cite{unforgettable_examples} for unforgettable examples which can be ignored without harming model's generalization. In addition, the stochastic gradient descent method tunes connective weights only in \textbf{a single epoch} where the salient feature lies in the idea of soft-forgetting. That is, the soft-forgetting technique crafts unique learning rate of each layer based on relevance of hidden layers to current concept safeguarding hidden layer parameters against external perturbation. 
	
	\begin{figure*}[t!]
		\begin{centering}
			\includegraphics[scale=0.63]{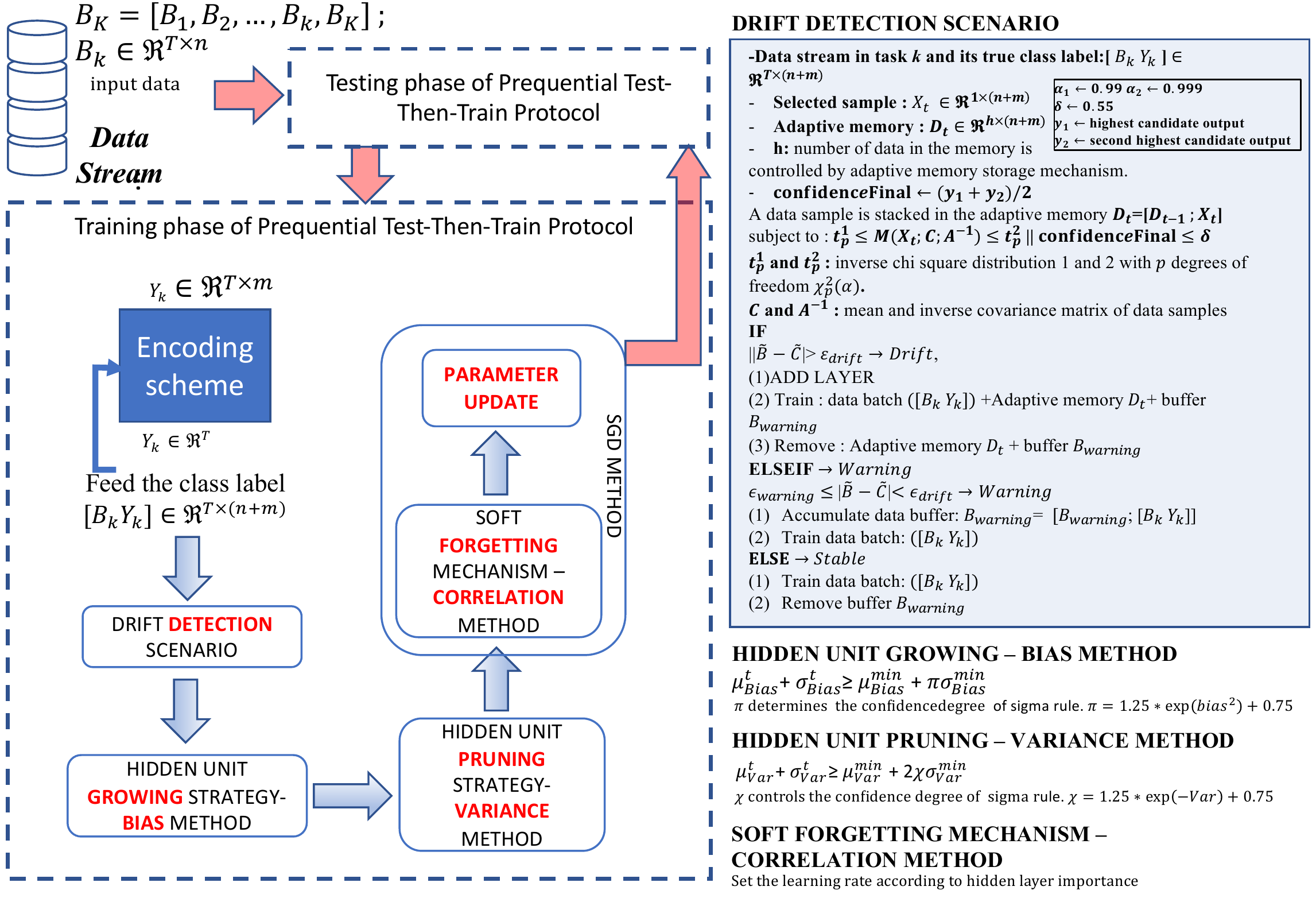}
			\par\end{centering}
		\caption{Overview of NADINE Learning Policy}
		\label{fig:PHO}
	\end{figure*}
	
	\section{Learning Policy of NADINE}
	This section elaborates on the NADINE's algorithm. Fig. 1 exhibits overview of NADINE algorithm and Fig. 2 exemplifies the structural evolution of NADINE. The pseudo-code of NADINE is presented in the supplemental document.
	
	\subsection{Adaptive Learning Strategy of Network Width}
	\textit{Hidden Unit Growing Strategy}: The hidden unit growing strategy is governed by the network significance (NS) method which quantifies the generalization power of network structure in terms of its statistical contribution under a certain probability density function. This aspect is formulated as an expectation of squared system error with a probability density function $p(x)$ as follows:
	\begin{equation}
		NS=\int_{-\infty}^{\infty}{(y-\tilde{y})}^2 p(x)dx
	\end{equation}
	Several mathematical derivations lead to the popular bias-variance formula as follows:
	\begin{equation}
		NS=E[(\tilde{y}-E[\tilde{y}])^{2}]+(E[\tilde{y}]-y)^{2}=Var(\tilde{y})+Bias(\tilde{y})^{2}    
	\end{equation}
	where the variance can be expanded further as $Var(\tilde{y})=E[\tilde{y}^{2}]-E[\tilde{y}]^{2}$. The key in solving the bias and variance formula is the solution of $E[\tilde{y}]=W_{out}\int_{-\infty}^{\infty}s(x W_D+b_D)p(x)dx+c_{out}$. It is assumed that $x$ is drawn from the normal distribution $p(x)=\frac{1}{\sqrt{2\pi\sigma}}exp(\frac{(x-\mu)^{2}}{\sigma^{2}})$ where $\mu,\sigma$ denote the empirical mean and standard deviation of data streams. It is known that the sigmoid function can be approached by a probit function $\Phi(\xi x)$ where $\Phi(x)=\int_{-inf}^{x}\aleph(\theta|0,1)d\theta$ and $\xi = \frac{\pi}{8}$. Therefore, $E[\tilde{y}]=W_{out}\int_{-\infty}^{\infty}\Phi(\xi x)p(x)dx+c_{out}$ can be derived by following the result in \cite{Murphy_Machine_Learning} where the integral of probit function is another probit function as follows:
	\begin{equation}\label{Expectation_Y}
		E[\tilde{y}]=W_{out}s(\frac{\mu}{\sqrt{1+\pi\frac{\sigma^{2}}{8}}}W_D+b_D)+c_{out}
	\end{equation}
	where $W_{out}\in\Re^{m \times R_D},b_{out}\in\Re^{m}$ stand for the connective weight vector and bias of the softmax layer. The bias expression can be established by plugging in (\ref{Expectation_Y}) to the bias term $Bias^{2}=(E[\tilde{y}]-y)^{2}$. This expression can be generalized to a deep network with the depth of $D$ by simply undertaking the multiple nonlinear mapping of sigmoid function $D$ times - forward-chaining. The bias formula approximates the predictive quality of DNN in respect to the probability density function, $p(x)$. A high bias signifies the underfitting situation which can be addressed by augmenting the network's complexity.  
	
	The hidden unit growing condition is designed by adopting the concept of statistical process control \cite{Gama2006} derived from the notion of $k$ sigma rule. The key difference lies in the calculation of mean and variance directly obtained from the bias itself rather than from the binomial distribution because the hidden unit growing strategy analyzes a real variable - bias instead of the accuracy score. The high bias problem leading to the introduction of new hidden unit is formulated as follows:
	\begin{equation}\label{HUgrowing}
		\mu_{bias}^{t}+\sigma_{bias}^{t} \geq \mu_{bias}^{min}+\pi\sigma_{bias}^{min}    \end{equation} 
	where $\pi=1.25exp(-Bias^{2})+0.75$ and controls the confidence degree of sigma rule. It is observed that $\pi$ is set adaptive as a factor of $Bias$ and revolves around $[1,2]$. A high bias results in low confidence level - close to $1$ attaining around $68.2\%$ confidence degree. On the other hand, a low bias renders high confidence level - close to $2$, equivalent to $95.2\%$. This method enhances flexibility of the hidden unit growing strategy because it adapts to variations of data streams and addresses the requirement of problem-specific threshold. $\mu_{bias}^{t},\sigma_{bias}^{t}$ stand for the empirical mean and standard deviation of bias up to $t-th$ time instant while $\mu_{bias}^{min},\sigma_{bias}^{min}$ denote the minimum bias up to $t-th$ time instant but is reset provided that (\ref{HUgrowing}) is satisfied. (\ref{HUgrowing}) is inherent to the existence of changing data distributions illustrated by the increase of network bias. The network bias is supposed to decrease or at least to be stable given that no drift occurs in data streams. 
	
	Once (\ref{HUgrowing}) is met, a new hidden unit is appended and its parameter $W_{R_D+1}$ is crafted as a negative system error $-e$. This strategy comes into picture to drive the system error toward zero as addition of new hidden unit while $b_{R_D+1}$ is assigned at random in the range of $[-1,1]$.
	
	\textit{Hidden Unit Pruning Strategy}: the hidden unit pruning strategy is crafted by following the similar principle as the hidden unit growing module but is based on the network variance instead of the network bias. If NADINE suffers from high variance, overfitting, the network complexity should be mitigated by alleviating the network complexity. The network variance can be modelled by first deriving the expression of $E[\tilde{y^{2}}]$ and $E[\tilde{y}]^{2}$. The later can be found by simply applying a squared operation to (\ref{Expectation_Y}) while the former can be expressed as $E[\tilde{y^{2}}]=W_{out} E[\Phi(\xi x)^{2}]+c_{out}$. Assuming that $y^{2}=y*y$ is the IID variable, the former is formulated as $E[\tilde{y^{2}}]=W_{out} E[\Phi(\xi x)]E[\Phi(\xi x)]+c_{out}$. The network variance is defined $Var(\tilde{y})=W_{out} E[\Phi(\xi x)]E[\Phi(\xi x)]-c_{out}+(W_{out}E[\Phi(\xi x)]+c_{out})^{2}$ where $E[\Phi(\xi x)]$ can be derived in the same manner as the bias. 
	
	The hidden unit pruning condition is set as the growing part derived from the statistical process control concept \cite{GamaDataStream} as follows:
	\begin{equation}\label{HUprune}
		\mu_{var}^{t}+\sigma_{var}^{t} \geq \mu_{var}^{min}+2\chi\sigma_{var}^{min}    
	\end{equation} 
	Compared to (\ref{HUgrowing}), the term $2$ is introduced and meant to avoid a direct pruning after adding situation since addition of a new hidden unit leads to temporary increase of network variance but gradually decreases as next observations are come across. $\chi$ is set akin to $\pi$ in the (\ref{HUgrowing}) as $\pi=1.25exp(-Variance^{2})+0.75$ which consequently causes $k$ sigma rule in the range of $[1,4]$. This strategy navigates to between 68.2\% and 99.9\% confidence level. 
	
	Once (\ref{HUprune}) is satisfied, the next step is about the complexity reduction strategy getting rid of the weakest component. This approach is undertaken by first measuring the contribution of every hidden unit by enumerating its statistical contribution. Assuming that the normal distribution assumption holds, the statistical contribution of a hidden unit is formalized from the expectation of activation function approached by the probit function as follows:
	\begin{equation}\label{UnitPruning}
		E[\Phi(\xi x)]=s(\frac{\mu}{\sqrt{1+\pi\frac{\sigma^{2}}{8}}}W_D+b_D)
	\end{equation}
	The pruning strategy is applied to the least contributing unit determined by (\ref{UnitPruning}) - having the lowest statistical contribution $Pruning\rightarrow\min_{i=1,...,R_D}E[\Phi(\xi x)]$. That is, the capacity of $D-th$ hidden layer shrinks $R_D=R_D-1$ as an attempt to relieve the overfitting situation. Moreover, the hidden unit growing and pruning strategy takes place in the forward-pass fashion at the top layer without looking back to preceding layers. This strategy is in line with the standard practise of transfer learning of DNN where only the last layer is trained using the target domain samples. Despite that, the calculation of expected output (\ref{Expectation_Y}) stems from the first layer being forward propagated to the last layer. Since a stochastic depth concept is implemented in NADINE where the depth of network structure is learned in a fully data-driven manner, the adaptive learning strategy of network width only takes place at the last layer. 
	
	\begin{figure*}[t!]
		\begin{centering}
			\includegraphics[scale=0.6]{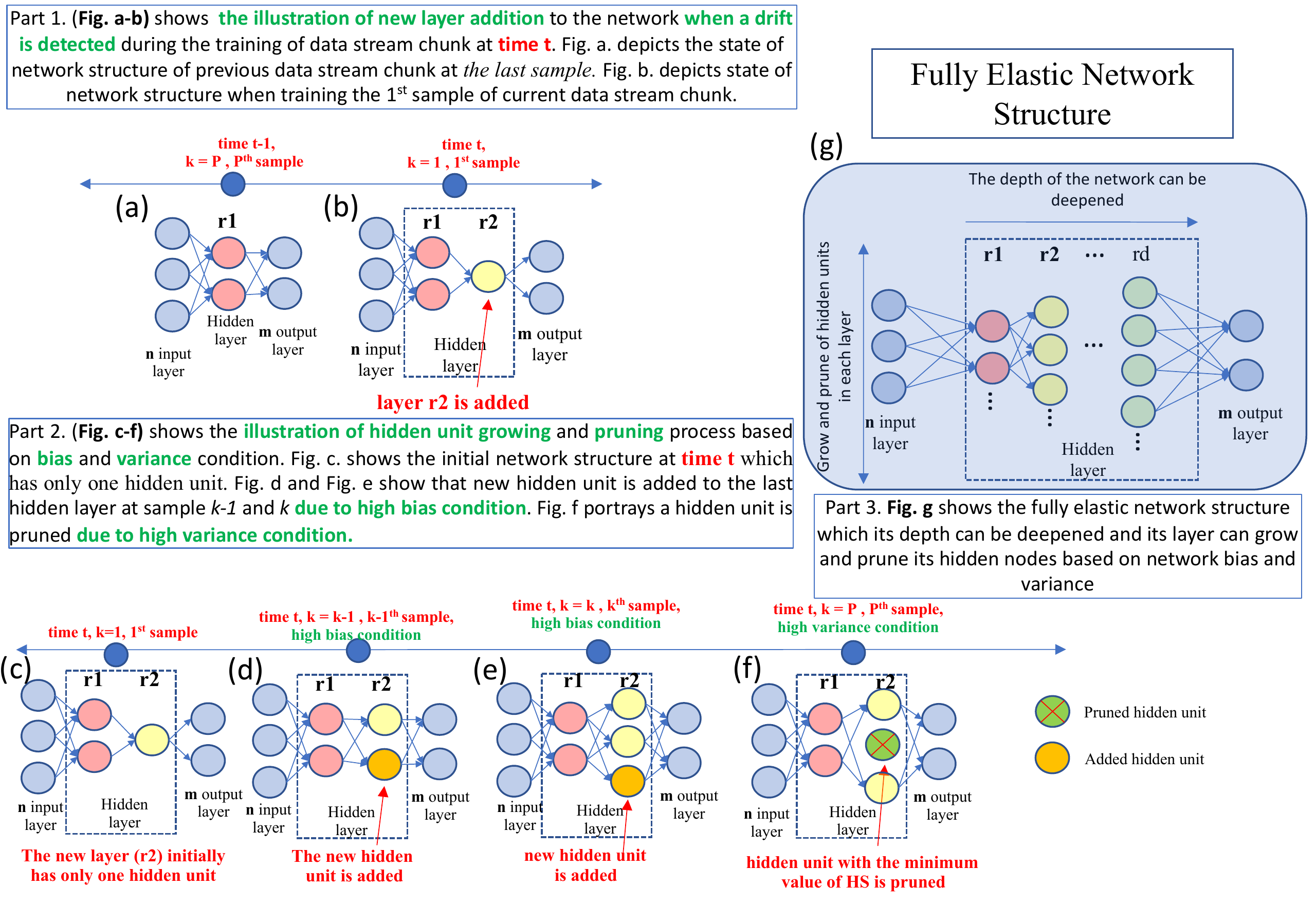}
			\par\end{centering}
		\caption{NADINE's structural evolution}
		\label{fig:PHO1}
	\end{figure*}
	
	\subsection{Adaptive Learning Strategy of Network Depth}
	The depth of NADINE actualizes a stochastic depth property governed by the drift detection method. A new hidden layer is incorporated if a concept drift is identified in the data streams. This strategy aims to increase the level of abstraction when changing training patterns are observed. That is, it links the representational level of data streams with the number of distinct concepts in data streams. Addition of depth improves generalization power because it increases the number of active regions \cite{linearregion}. It is understood from \cite{powerofdepth} that DNN is proven theoretically to be more powerful than a shallow neural network simply because it conveys a higher number of local regions. Although the concept drift detection method has been widely used as a knowledge-base expansion strategy of ensemble classifier \cite{pENsemble}, the ensemble classifier has loosely coupled connections among their components - no cross correlation across different components and has little risk of the catastrophic forgetting problem due to its modular characteristic.
	
	To the best of our knowledge, this study is the first attempt to utilize the drift detection scenario for the hidden layer growing mechanism in the standard multi-layer structure. The drift detection mechanism enhances the prominent Hoeffding's bound detection method \cite{Pesaranghader2018} integrating an adaptive windowing strategy. The adaptive windowing strategy puts forward a flexible predictive horizon of drifts using the idea of switching points. That is, the switching point here refers to the increase of population mean. An accuracy matrix is constructed and records predictive performance of NADINE during the testing phase of prequential-then-train procedure. This strategy enables the identification of real drift, variations of input domain leading to the changes of target domain. Note that the real drift is more severe than the covariate drift since it compromises the predictive performance \cite{Gamaconceptdrift} due to the drifting decision boundary. In addition, the Hoeffding's bound method is general for any type of data distribution.  
	
	The accuracy matrix $A\in\Re^{T}$ records the generalization performance of the testing phase. $1$ is returned given that misclassification happens $\tilde{y_t}\neq y_t$  while $0$ is inserted if NADINE predicts correctly a testing sample $\tilde{y_t}= y_t$. The switching point is found by evaluating two partitions of accuracy matrix $A\in\Re^{T},B\in\Re^{cut}$ where $cut$ is the hypothetical switching point of interest. A switching point is discovered in respect to the following condition. 
	\begin{equation}\label{switching}
		\tilde{A}+\epsilon_{\tilde{A}}\leq\tilde{B}+\epsilon_{\tilde{B}}
	\end{equation}
	where $\tilde{A},\tilde{B}$ are the statistics of accuracy matrix $A,B$ while $\epsilon_{\tilde{A}},\epsilon_{\tilde{B}}$ are their corresponding Hoeffding's error bounds. Furthermore, every sample is treated equally excluding any weights to provide timely response to sudden change. The Hoeffding's error bounds are calculated as follows:
	\begin{equation}\label{hoeffding}
		\epsilon=(b-a)\sqrt{\frac{size}{2*size*cut}ln{\frac{1}{\alpha}}}
	\end{equation}
	where $size$ denotes the size of accuracy matrix $A,B$ and $\alpha$ stands for the significance level of Hoeffding bound. Note that $\alpha$ here is inversely proportional to the confidence level $1-\alpha$ and statistically justifiable. It is not categorized as a case-sensitive hyper-parameter because the lower the value the higher the confidence level is returned - goes to 100\% while the higher the value the lower the confidence level is produced - approaches 0\%. Furthermore, $\epsilon_{\tilde{A}},\epsilon_{\tilde{B}}$ in (\ref{switching}) are obtained using (\ref{hoeffding}) by applying $\alpha_{\tilde{A}},\alpha_{\tilde{B}}$ where both of them are set to be equal. Note that $\alpha_{\tilde{A}},\alpha_{\tilde{B}}$ should not be seen as problem-specific hyper-parameters because they have clear statistical interpretation. 
	
	The partition of accuracy matrix $C\in\Re^{T-cut}$ can be formed once eliciting the cutting point $cut$. This partition of accuracy matrix will be used as a reference whether or not null hypothesis is violated. The null hypothesis examines the increase of statistics of accuracy matrix which confirms the concept drift situation. That is, $\tilde{B},\tilde{C}$ are compared $|\tilde{B}-\tilde{C}|$ and null hypothesis is rejected provided that the gap between the two exceeds $|\tilde{B}-\tilde{C}|\geq\epsilon_{drift}$. $\epsilon_{drift}$ is the Hoeffding's bound sharing the same characteristic as $\epsilon_{\tilde{A}},\epsilon_{\tilde{B}}$. Another condition, namely warning, is put in place and points out a transition situation where further observation is required to signal a concept drift - the gradual drift. This condition is formalized by introducing $\epsilon_{warning}$ where it has a lower confidence level than the drift situation $\epsilon_{drift}>\epsilon_{warning}$. Formally speaking, the warning situation is indicated by $\epsilon_{warning}\leq |\tilde{B}-\tilde{C}|<\epsilon_{drift}$. The stable condition is returned if data streams violate both conditions. Moreover, both $\epsilon_{drift},\epsilon_{warning}$ are set in similar way as $\epsilon_{\tilde{A}},\epsilon_{\tilde{B}}$ by applying their specific significance levels $\alpha_{drift},\alpha_{warning}$ to (\ref{hoeffding}). Because the warning phase has to be less confident than the drift phase $\epsilon_{drift}>\epsilon_{warning}$, their significance level should satisfy  $\alpha_{drift}<\alpha_{warning}$. In addition, we adopt $\alpha_{drift}=\alpha_{\tilde{A}}=\alpha_{\tilde{B}}$ because they are used to signify the switching points. As with $\alpha_{\tilde{A}},\alpha_{\tilde{B}}$, $\alpha_{drift},\alpha_{warning}$ have clear statistical meaning and are thus not classified as problem-specific parameters. 
	
	The increase of empirical mean portrays a strong indication of concept drift because classification error is expected to be stable or decreases in the drift-free situation. That is, $|\tilde{B}-\tilde{C}|\geq\epsilon_{drift}$ presents a case where the empirical mean of $B$ is lower than the empirical mean of data batch $A$ reflecting the fact where the classification performance deteriorates. The drift case triggers the hidden layer insertion procedure deepening the depth of network structure. An initialization phase of new hidden layer has to be carefully designed because it paves a way to the catastrophic forgetting problem due to the standard multi-layer network architecture where the final output is fully produced by the end layer. The adaptive memory strategy is proposed to resolve this situation where it memorizes key data points which supports perfect reconstruction of overall data distribution. This procedure is detailed in the next section. During the warning phase, data batch $B_k$ is accumulated in the buffer $B_{warning}=[B_{warning};B_k]$ and is used to induce a new layer if the drift condition is alarmed. The stable condition incurs refinement of current structure by means of the one-pass SGD method. The innovation of NADINE lies in the proposal of soft forgetting which crafts unique learning rates per layer in respect to hidden layer relevance.   
	
	\subsection{Solution of Catastrophic Forgetting}
	{\textit{Adaptive Memory Strategy}:} the concept of adaptive memory refers to selective sampling procedure to discover important data points revealing the structure of problem space. Adaptive memory memorizes important data points to be replayed during insertion of a new hidden layer which overcome loss of generalization power due to the catastrophic forgetting of previously valid knowledge. The adaptive memory is developed from the concept of ellipsoidal anomaly detector in \cite{masud} covering only the issue of anomaly detection. It is generalized here for selecting key samples to be replayed if the condition of hidden layer addition holds. Moreover, important sample is characterized in two conditions: 1) it must represent the underlying orientation of previously seen samples but must not be redundant. Sample redundancy is a vital issue here because redundant samples cause the over-fitting situation. Redundant sample can be also perceived as unforgettable example which can be remembered forever once learned \cite{unforgettable_examples}. Moreover, the adaptive window should be kept in modest size to ensure NADINE's scalability in the resource-constrained environment. 2) the adaptive memory must avoid any outliers to reject low variance direction of data distribution. 
	
	Suppose that $C\in\Re^{n},A^{-1}\in\Re^{n \times n}$ are the mean and inverse covariance matrix of data samples calculated recursively as in \cite{masud}, the Mahalanobis distance is used to evaluate sample's importance $M(X_t;C,A^{-1})$. A data sample is stacked in the adaptive memory $D_t=[D_{t-1};X_t]$ subject to:
	\begin{equation}\label{anomaly}
		t_p^{1} \leq M(X_t;C,A^{-1}) \leq t_p^{2}
	\end{equation}
	where $t_p^{1},t_p^{2}$ are selected according to the inverse of the chi-square distribution with $p$ degrees of freedom $\chi_p^{2}(\alpha)$. $p$ is set as the input dimension and $\alpha$ denotes the confidence level. Since region of interest is created using two thresholds, the condition $\alpha_1 < \alpha_2$ is established and respectively assigned as $0.99,0.999$. (\ref{anomaly}) implies the normal distribution assumption is enforced. This facet should not distract the efficacy of adaptive memory since the threshold selection from any unimodal distribution covers the majority of data points \cite{masud}. In addition, $\alpha$ is chosen  above $0.99$ leading to $99\%$ confidence level. The adaptive memory is created to realize the experience-replay mechanism. That is, all samples of the adaptive memory are replayed if a new hidden layer is added. 
	
	Notwithstanding that it is capable of observing abnormal samples, (\ref{anomaly}) returns too few samples because of the uniqueness of anomalous samples and does not sufficiently tell previous system dynamics but should be included to disclose hidden characteristics of system dynamic. In addition, hard examples are sampled in the adaptive memory and are defined as those to which a model is not capable of predicting with high confidence. Such samples occupy an adjacent area of decision boundary. A sample is integrated in the adaptive memory subject to the following condition.
	\begin{equation}\label{confidence}
		\frac{y_1}{y_1+y_2}<\delta
	\end{equation}
	where $y_1,y_2$ respectively denote the highest and second highest outputs of softmax layers and $\delta$ is a user-defined threshold. Note that the output of softmax layer has a clear probabilistic interpretation. Hence, a classifier confusion is obvious when the ratio of its highest and second highest outputs is minor. This situation consequently causes $\frac{y_1}{y_1+y_2}$ to be close 0.5. This fact explains the fact that $\delta$ is not problem-specific and is fixed at $0.55$ for all our simulations.  
	
	{\textit{Soft Forgetting Strategy}:} the soft forgetting strategy is applied in NADINE and controls the rate of parameter changes of hidden layers in respect to their relevance to target outputs. This strategy is devised to suit the hidden layer growing strategy utilizing the concept drift detection method. That is, every layer is constructed in different time intervals portraying different concepts of data streams. This strategy impedes hidden layer parameters to move from their designated zones. 
	
	The relevance of hidden layer to the target classes can be quantified by their correlation using the Pearson correlation method as follows:
	\begin{equation}
		Corr(H_d^{i_d},Y_o)=\frac{Cov(H_d^{i_d},Y_o)}{\sigma_{H_d}^{i_d}\sigma_{Y_o}}    
	\end{equation}
	where $Cov(H_d^{i_d},Y_o),\sigma_{H_d}^{i_d}\sigma_{Y_o}$ are the correlation of the $i_d$ hidden node of $d-th$ hidden layer and $Y_o$ target variable, the standard deviation of the $i_d$ hidden node of $d-th$ hidden layer and the standard deviation of $Y_o$ target variable respectively. Note that the Pearson correlation coefficient is computed per data batch $B_k$. The relevance score of $d-th$ hidden layer, $RS_d$, is defined as the average correlation of hidden units at the $d-th$ hidden layer $RS_d$. The learning rate of $d-th$ hidden layer is produced:
	\begin{equation}\label{LearningRate}
		\eta_d=0.01exp(-1/RS_d-1)
	\end{equation}
	where $\eta_d$ is the hidden layer-specific learning rate to be applied in the SGD process. $\eta_d$ is bounded in the range of $[0,0.01]$ where the higher the correlation to the target class the higher the learning rate is allocated. This strategy enables relevant hidden layers to adapt to variations of data streams while irrelevant layers are frozen from being distracted to unsuitable concepts.

	\section{Proof of Concepts}
	\noindent\textbf{Experimental Setting}: this section outlines numerical validation of NADINE using six popular data stream problems: SEA \cite{DitzlerImbalanced}, Susy, HEPMASS \cite{Baldi2014SearchingFE}, KDDCup \cite{KDDCup}. These problems feature non-stationary properties to examine the evolving characteristics of NADINE or are big in size making possible to simulate lifelong learning environments. In addition, permutted MNIST \cite{permuttedMNIST}, rotated MNIST \cite{rotatedMNIST}, are included to demonstrate NADINE's aptitude to handle non-stationary image classification problems having high input dimension. Table \ref{tab:DatasetProperties} tabulates the properties of six problems. The detailed descriptions of the six problems are made available in the supplemental document.
	\begin{table}[htbp]
		\caption{Properties of dataset}
		\begin{centering}
			\scalebox{0.8}{
				\begin{tabular}{ccccccc}
					\toprule
					Data Set & $\#$IA & C & Samples & Folds & Characteristics & Task\tabularnewline
					\midrule
					SUSY & 18 & 2 & 5M & 5K & stationary & Classification\tabularnewline
					HEPMASS 19\% & 28 & 2 & 2M & 2K & stationary & Classification\tabularnewline
					Household Elec & 4 & 3 & 2049280 & 2049280 & Non-stationary & Regression\tabularnewline
					SP500 & 5 & 1 & 29786 & 29786 & Non-stationary & Regression\tabularnewline
					Cond Monitoring & 16    &  2     &  11934   & 11934     &Non-Stationary & Regression\tabularnewline
					Rotated MNIST & 784    &  10     &  65K   & 65     &Non-stationary & Classification\tabularnewline
					Permuted MNIST & 784    &  10     &  70K   & 70     &Non-stationary & Classification\tabularnewline

					KDDCup 10\% & 41 & 2 & 500K & 500 & Non-stationary & Classification\tabularnewline
					SEA & 3 & 2 & 200K & 200 & Non-stationary & Classification\tabularnewline
					\bottomrule
			\end{tabular}}
			\par\end{centering}
		\centering{}IA: input attributes, C: classes, Sample: number of data points
		\label{tab:DatasetProperties}%
	\end{table}	
	
	Our numerical study is carried out via the prequential test-then-train procedure \cite{GamaDataStream} where the testing phase is run before the training process and this procedure is carried out continually over a number of tasks $K$. This scenario mirrors the fact that data stream arrives without labels and a delay is expected in labelling data points. Numerical results are obtained from the average of numerical results across all time stamps. Furthermore, the number of tasks is set large while a small number of samples per task is assigned. The source code of NADINE can be found from \url{https://www.researchgate.net/publication/335813786_NADINE_Code_M_File} to assure reproducible results. Numerical results of consolidated algorithms are reported in Table \ref{tab:NumResult1} and are taken from the average of five consecutive runs. NA indicates the computational constraint where the numerical results cannot be obtained after running the algorithm for days. 
	
	\noindent\textbf{Baselines}: NADINE is compared against seven prominent data stream algorithms: progressive neural networks (PNN) \cite{progressiveneuralnetworks}, dynamically expandable network (DEN) \cite{DeepExpandable}, hard attention to the task \cite{hardattention}, incremental bagging and boosting \cite{IncBaggingBoosting}, autonomous deep learning (ADL) \cite{ADL}, online multiclass boosting \cite{OnlineMBM}. All consolidated algorithms are executed under the same computational platforms using their published codes. Numerical results of consolidated algorithms are presented in Table \ref{tab:NumResult1}. 
	
	\noindent\textbf{Setting of Hyper-parameters}: The significance levels of NADINE,   $\alpha_{drift},\alpha_{\tilde{A}},\alpha_{\tilde{B}}$ are set at $0.0001$ to return $99.99\%$ confidence level while $0.0005$ is chosen for $\alpha_{warning}$ thus inducing $99.95\%$ confidence level. The confidence threshold $\delta$ is set at 0.55 which corresponds to 55\% of confidence level. These values are kept fixed across all simulations in this paper to demonstrate the non ad-hoc property of NADINE. On the other hand, all predefined parameters of other benchmarked algorithms are hand-tuned and the best-performing results of many combinations are reported here. It is worth mentioning that we obtain significant performance degradation of PNN, HAT and DEN when assigning them more complex network structure than at the current level. Their current network structure here is the best-performing configuration among many others in our experiments. 
	
	\noindent\textbf{Numerical Results}:The advantage of NADINE is clear from Table \ref{tab:NumResult1} in which it produces the most encouraging numerical results in all six problems. The advantage of NADINE is obvious in the permutted MNIST problem where the performance gap to the second best performing algorithm is close to $10\%$. It is worth stressing that numerical results of Incremental Bagging and Incremental Boosting in the three problems, namely PermuttedMNIST, RotatedMNIST and SUSY, are not reported here because the two approaches do not scale with the large-scale problems having high dimension of input space or data space.  Moreover, the incremental boosting do not support the multi-class classification problem without the application of one versus rest classification scheme. It is also found in our numerical study that a model selection plays key role for the successful application of PNN, DEN, HAT and significant amount of efforts are paid here to select their best network structure. This fact bears out the advantage of NADINE where the learning process can be automated without over-dependence on problem specific hyper-parameters. Moreover, all of which are trained in the one-pass training fashion as NADINE. Another interesting observation is present in the varying network capacities of NADINE across different problems. This aspect substantiates the open structure of NADINE where its structure is fully data driven and self-evolved from scratch. Although ADL also possesses an elastic network width and depth, it is framed under the different-depth structure having local softmax layer and hindering ADL's application to the non-classification task. In realm of network complexity, NADINE evolves more compact structure than ADL due to the absence of local softmax layer. NADINE's runtime is also comparable where it is consistently faster than PNN, HAT and DEN while being in the same level as those incremental boosting and ADL. 
	\begin{table*}[!t]
		\caption{Numerical results of benchmarked algorithms}
		\begin{center}
			\scalebox{0.7}{
				\begin{tabular}{clrrrrr}
					\toprule
					Data Set & \multicolumn{1}{c}{Methods} & \begin{tabular}   {@{}c@{}}Classification \\ rate (\%)\end{tabular} & \begin{tabular}   {@{}c@{}}Execution \\ time(s)\end{tabular} & \begin{tabular}   {@{}c@{}}Hidden \\ Layers\end{tabular} & \begin{tabular}   {@{}c@{}}Hidden \\ Nodes\end{tabular} & \begin{tabular}   {@{}c@{}}Number of \\ parameters\end{tabular}\\
					\midrule
					& NADINE & 	\textbf{78.03	$\pm$	3}	&	1455	&	4.10	$\pm$	1.87	&	111.88	$\pm$	65.01	&	(	3.07	$\pm$	2.24	)K\\
					& DEN & 63.15 $\pm$ 10.06 & NA & 1 $\pm$ 0 & 10 $\pm$ 0 & 212 $\pm$ 0 \\
					& HAT    & 73.85 $\pm$ 3.18 & NA  & 2 $\pm$ 0     & 20 $\pm$ 0 & 342 $\pm$ 0 \\
					SUSY & Incremental Bagging    &     72.8   $\pm$   3.1   &   73K   &     NA     &      NA     &       NA      \\
					& Incremental Boosting &  72.8   $\pm$  1.4    & 374     &     NA     &      NA     &       NA      \\
					& PNN & 67.75 $\pm$ 3.71 & NA  & 3 $\pm$ 0 & 22 $\pm$ 0 & 424 $\pm$ 0\\
					& ADL &  77.95$\pm$3.15  & 1.24K  &  2.28$\pm$1.38 &  186.21$\pm$25.04  &  1.82K$\pm$2.03K  \\
					& Online Multiclass Boosting &  77.73$\pm$1.43  & 14.4K  &  NA &  NA  &  NA \\
					\midrule
					& NADINE & \textbf{83.82	$\pm$	2}	&	499	&	3.39	$\pm$	1.71	&	67.90	$\pm$	43.33	&	(	1.49	$\pm$	1.01	)K	\\
					& DEN &  67.25 $\pm$ 15.05 & NA & 1 $\pm$  0  & 14 $\pm$ 0   & 436 $\pm$ 0 \\
					& HAT    & 74.62 $\pm$ 4.11 & NA  & 2 $\pm$ 0     & 40 $\pm$ 0 & 1.1K $\pm$ 0 \\
					HEPMASS 19\% & Incremental Bagging    &   78.3     $\pm$  2.1    &  2.3K    &     NA    &      NA     &       NA      \\
					& Incremental Boosting &   80.1  $\pm$  1.3    &  60    &     NA    &      NA     &       NA      \\
					& PNN & 78.31 $\pm$ 3.07 & NA  & 3 $\pm$ 0 & 51 $\pm$ 0 & 1.4K $\pm$ 0 \\
					& ADL &  83.65$\pm$ 3  & 476  &  2.75$\pm$1.98 &  47.19$\pm$46.14  &  940$\pm$984  \\
					& Online Multiclass Boosting &  83.12$\pm$1.23  & 17.05K  &  $\pm$  &  $\pm$  &  $\pm$ 0 \\
					\midrule
					
					& NADINE & \textbf{	74.51	$\pm$	7.50}	&	192	&	1	$\pm$	0	&	15.69	$\pm$	4.188	&	(	12.67	$\pm$	2.944	)K	\\
					& DEN & 61.48$\pm$21.75  & NA  & 2$\pm$0 & 440 &  290K  \\
					& HAT &  54 $\pm$ 8.77  & NA  &  2 $\pm$ 0  &  60 $\pm$ 0  & 24.9K $\pm$ 0     \\
					Rotated Mnist& Incremental Bagging    &        NA      & NA     &     NA     &      NA     &       NA      \\
					& Incremental Boosting &     NA      &  NA    &     NA     &      NA     &       NA      \\
					& PNN &60.94  $\pm$11.25  & NA  & 3 $\pm$ 0 & 750 $\pm$ 0 & 503K $\pm$ 0\\
					& ADL &  72.90$\pm$9.35  & 199  &  1.90$\pm$0.29 &  8.73$\pm$1.16  &  7.4K$\pm$600  \\
					& Online Multiclass Boosting &  26.07$\pm$5.8  & 5.3K  &  NA  &  NA  &  NA \\
					\midrule
					& NADINE & 	\textbf{77.65	$\pm$	15.09}	&	202	&	1.19	$\pm$	0.11	&	22.33	$\pm$	6.23	&	(	12.96	$\pm$	2.17	)K	\\
					& DEN & 52.08$\pm$22.6  & NA  & 2$\pm$0 & 440 $\pm$ 0 &  290K $\pm$ 0 \\
					& HAT &  63 $\pm$ 16.2  & NA  &  2 $\pm$ 0  &  60 $\pm$ 0  & 24.9K $\pm$ 0     \\
					Permuted Mnist & Incremental Bagging    &        NA      & NA     &     NA     &      NA     &       NA      \\
					& Incremental Boosting &     NA      &  NA    &     NA     &      NA     &       NA      \\
					& PNN & 64.42 $\pm$ 8.77 & NA  & 3 $\pm$ 0 & 705 $\pm$ 0 & 503K $\pm$ 0\\
					& ADL &  68.40$\pm$24.17  & 212  &  1.39$\pm$0.49 &  19.81$\pm$4.26  &  16.2K$\pm$3.01K  \\
					& Online Multiclass Boosting &  11.22$\pm$0.97  & 7.1K  &  NA  &  NA  &  NA \\
					\midrule
					& NADINE & \textbf{99.84	$\pm$	0.15}	&	98	&	1	$\pm$	0	&	12.33	$\pm$	0.57	&		545.41	$\pm$	10.90	\\
					& DEN & 99.73 $\pm$ 0.19 &  NA & 1 $\pm$ 0  & 20 $\pm$  0  & 860 $\pm$ 0 \\
					& HAT &  99.51$\pm$ 3.41& NA  &  2 $\pm$ 0   & 60 $\pm$ 0   &2.3K $\pm$ 0   \\
					KDD Cup 10\% & Incremental Bagging    &   99.5     $\pm$  0.4    &   1.6K   &     NA     &      NA     &       NA      \\
					& Incremental Boosting &  98.55   $\pm$   0.4   &  4.89    &     NA     &     NA    &       NA      \\
					& PNN & 94.71 $\pm$ 1.79 & NA  & 3 $\pm$ 0 & 375 $\pm$ 0 & 41.9K $\pm$ 0 \\
					& ADL &  99.84$\pm$0.15  & 115  &  1$\pm$0 &  12.68$\pm$0.73  &  561.36$\pm$22.50  \\
					& Online Multiclass Boosting & 99.64 $\pm$ 0.23  & 1.48K  &  $\pm$  &  $\pm$  &  $\pm$ 0 \\
					\midrule
					
					& NADINE & 	\textbf{92.24	$\pm$	6.40}	&	15	&	1.19	$\pm$	0.26	&	12.91	$\pm$	8.74	&	114.37	$\pm$	116.04	\\
					& DEN & 79.95 $\pm$ 19.28 & NA & 1 $\pm$ 0   & 6 $\pm$ 0  & 38 $\pm$ 0 \\
					& HAT    & 74.65 $\pm$10.1  & NA  & 2 $\pm$ 0     & 10 $\pm$ 0 & 72 $\pm$ 0 \\
					SEA & Incremental Bagging    &   84.6     $\pm$   13   &   1.5K   &     NA     &      NA     &       NA      \\
					& Incremental Boosting &  77.9   $\pm$  6.1    &  0.21    &     NA     &      NA     &       NA      \\
					& PNN & 75.27 $\pm$ 2.31 & NA  & 3 $\pm$ 0 & 33 $\pm$ 0 & 347 $\pm$ 0\\
					& ADL &  92.13$\pm$0.0906  & 14  &  1$\pm$0 &  11.42$\pm$2.79  &  71.18$\pm$15.58  \\
					& Online Multiclass Boosting &  87.86$\pm$3.85  & 77.4  &  NA  &  NA  &  NA \\
					\bottomrule
				\end{tabular}
			}%
		\end{center}
		
		\label{tab:NumResult1}%
	\end{table*}%
	
	The open structure property of NADINE is pictorially demonstrated in Fig. \ref{fig:fig2} where it is capable of starting its learning process from scratch with the absence of initial structure. Its hidden node and hidden layer are automatically grown and pruned in respect to model's generalization power. It is evident that a drop of classification performance is timely responded by structural evolution in which new node and layer are added if NADINE's learning performance is compromised. The structural evolution is initiated by growing and pruning hidden nodes up to a point where a drift detector signals possible concept change - addition of new layer. The advantage of stochastic depth feature is seen where it incurs significant improvement of learning performance. The classification rate gradually increases or is stable as the number of tasks while the network loss revolves around a small and bounded range.  
	\begin{figure}[t!]
		\begin{centering}
			\includegraphics[scale=0.6]{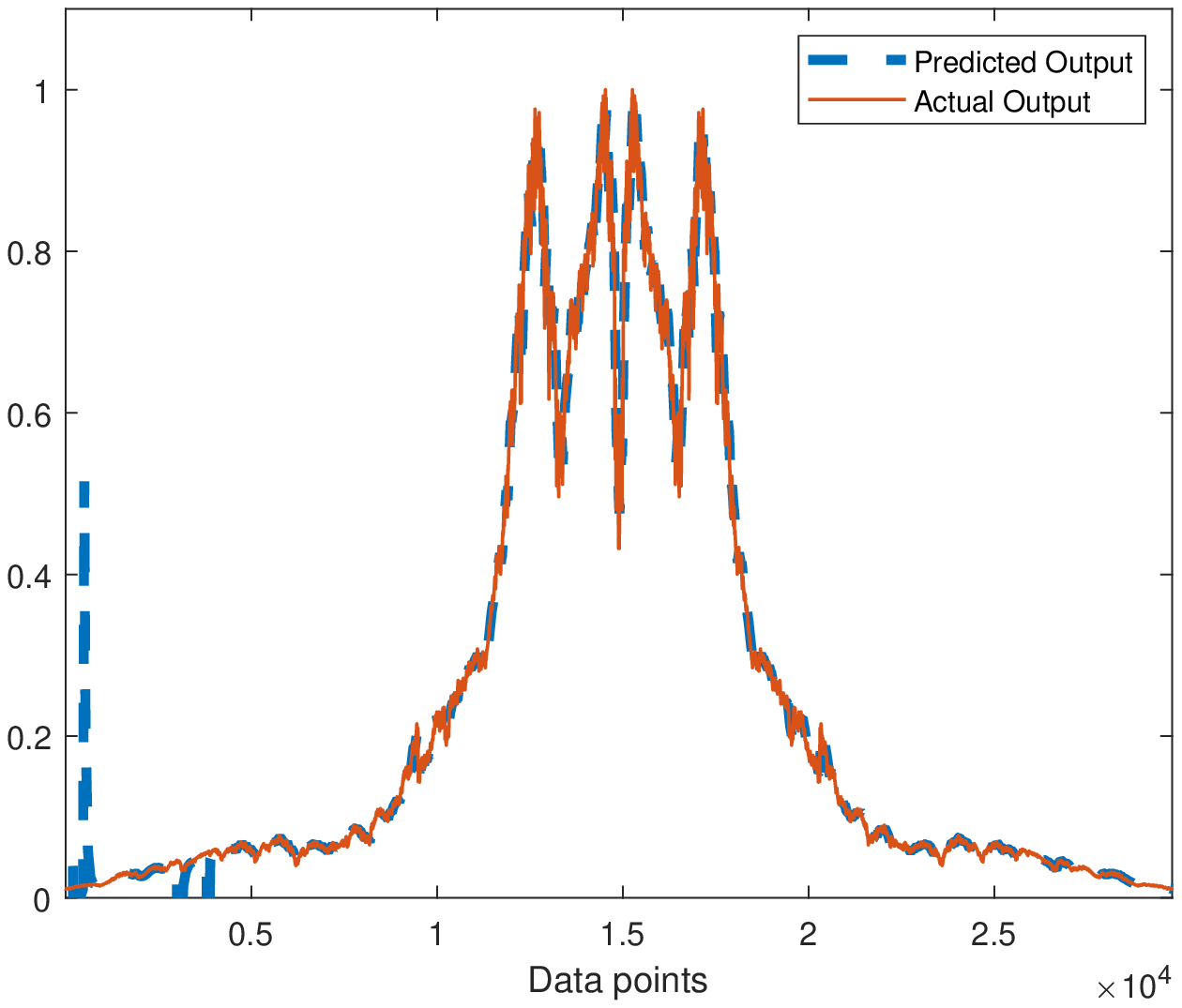}
			\par\end{centering}
		\caption{Actual and predicted outputs for SP500 data}
		\label{fig:SP500}
	\end{figure}
	
	\noindent\textbf{Regression Problems}: NADINE is capable of handling the regression problem equally well as the classification problem because it is developed under the MLP-like structure. NADINE's performance for regression problem is numerically validated in three data stream regression problems with non-stationary characteristics, household electricity prediction, SP 500 time series index prediction and condition monitoring problems \cite{ConditionMonitoring} where their characteristics are summarized in Table \ref{tab:DatasetProperties} and detailed in the supplemental document. Minor modification is performed in NADINE where the drift detection module checks the prequential error rather than classification error. Furthermore, the household electricity prediction problem and the tool condition monitoring problem presents the MIMO problems with three and two target variables respectively. Our simulation is performed using the prequential test-then-train protocol as the classification problem but the predictive accuracy is measured per sample rather than per data batch. NADINE is compared against DNN with a fixed structure where its network structure is set such that it has comparable network capacity as NADINE to ensure fair comparison. It is trained in the single pass mode using the SGD method as NADINE. 
	\begin{table*}[!t]
		\caption{Numerical results of NADINE in Regression Problems}
		\begin{center}

			\label{result-2-1-1} \scalebox{0.55}{ %
				\begin{tabular}{llrrrrrrr}
					\toprule
					& ALGORITHM & RMSE  & NDEI & TrT & TsT & HN & HL & NoP\tabularnewline
					\midrule
					SP500 & NADINE  & $0.0798$ & $0.2816$ & $0.0028\pm{0.0017}$  & $0.0009\pm{0.0002}$ & $34.97\pm{0.15}$  & $6.2610\pm{2.1641}$ & $187.9334\pm{115.2440}$ \tabularnewline
					& DNN & $0.0887$ & $0.3129$ & $0.0005\pm{0.00003}$  & $0.0002\pm{0.00002}$ & $35$  & $7$  & $240$ \tabularnewline
					\midrule 
					Household & NADINE  & $[0.0562;0.0502;0.1322]$ & $[0.8041;0.6897;0.4857]$ & $0.02\pm{0.03}$  & $0.08\pm{0.06}$ & $81\pm{0.019}$ & $6.81\pm{0.56}$ & $(1.2\pm{0.1})$K \tabularnewline
					Electric Power & DNN & $[0.0374;0.0340;0.0728]$ & $[0.5345;0.4678;0.2674]$ & $0.0007\pm{0.0001}$  & $0.0002\pm{0.00002}$ & $80$ & $5$ & $1$K \tabularnewline
					\midrule 
					Condition & NADINE  & $[0.1141;0.2974]$ & $[1.0830;0.7592]$ & $0.003\pm{0.0012}$  & $(6.6\pm{2.699})\times10^{-4}$  & $30.9770\pm{0.3494}$  & $1.9917\pm{0.0907}$  & $546.0655\pm{30.4762}$ \tabularnewline
					Monitoring & DNN & $[0.1889;0.2919]$ & $[1.7862;0.7451]$ & $0.0003\pm{0.0001}$  & $0.0003\pm{0.0001}$  & $15$  & $3$  & $400$ \tabularnewline
					\bottomrule
			\end{tabular}} 
		\end{center}
		\centering{}CR: classification rate, TrT: training time, TsT: testing
		time, HN: hidden nodes, HL: hidden layers, NoP: number of parameters 
		\label{Regression}
	\end{table*}
	
	It is demonstrated in Table \ref{Regression} that NADINE outperforms DNN in two problems, SP 500 and condition monitoring problems. It is slightly inferior to DNN in the household electricity consumption problem. The household electricity consumption problem, however, contains over 2 million records recorded over period of 4 years which sufficiently covers temporal nature of electricity usage. Moreover, it is only measured from one household such that it has predictable and regular characteristics. The most relevant type of drift in this problem is the cyclical drift as the daily electricity usage: peak and off-peak period. On the other hand, SP 500 has volatile characteristics as the nature of stock data. Furthermore, this problem also captures the period of financial recession in the US around 2008-2009. The condition monitoring problem concerns on the condition-based maintenance of naval propulsion taking into account the performance decay overtime - the gradual drift. In short, the advantage of NADINE over DNN is noticeable in the highly uncertain and non-stationary problems. Fig. \ref{fig:SP500} pictorially illustrates the predictive performance of NADINE in performing one day ahead time series prediction of SP500 daily index in the test-then-train fashion where it exhibits accurate behaviour. 
	
	\noindent\textbf{Ablation Study and Sensitivity Analysis}: Table
	\ref{tab:Tab1} exhibits the numerical results of NADINE under four different learning configurations and different confidence levels $\alpha_{drift},\alpha_{warning}$. Note that $\alpha_{\tilde{A}}=\alpha_{\tilde{B}}=\alpha_{drift}$. This study is carried out using the SEA problem. NADINE suffers from 3-7\% performance degradation with the deactivation of layer growing, node pruning, memory replay and soft forgetting confirming their notable contribution to the learning performance. It is also demonstrated that the application of hidden layer expansion strategy without either the adaptive memory and soft forgetting methods compromises the learning performance. The absence of hidden unit pruning mechanism causes the predictive performance to suffer while imposing the increase of network complexity. On the other hand, variation of confidence levels leads to minor performance difference which bears out their non problem-specific characteristics. 	\begin{table*}[!t]
		\caption{Ablation Study}
		
		\begin{center}
			\scalebox{0.8}{
				\begin{tabular}{p{7.57em}ccccccc}
					\toprule
					
					Type execution & Classification & Execution  & Hidden & Hidden & Number of & $\alpha_{drift}$ & $\alpha_{warning}$ \\
					& Rate  & Time (s) & Layers & Nodes & parameters &  &  \\
					\midrule
					Without  & 89.06$\pm$6.42  & 18    &  1.00$\pm$0.00  &  10.10$\pm$2.80  &  65.30$\pm$10.70  & 0.0003 & 0.0002 \\
					Layer Growing  &  88.83$\pm$6.42  & 13    &  1.00$\pm$0.00  &  9.60$\pm$3.40  &  62.70$\pm$16.40  & 0.0007 & 0.0002 \\
					Mechanism & 85.85$\pm$10.41 & 14    &  1.00$\pm$0.00  &  5.20$\pm$3.70  &  34.30$\pm$22.10  & 0.0005 & 0.0001 \\
					\midrule
					Without  &  89.85$\pm$6.32  & 14    &  1.06$\pm$0.24  &  14.70$\pm$8.10  &  114.30$\pm$127.20  & 0.0003 & 0.0002 \\
					Hidden Nodes &  87.23$\pm$8.63  & 15    &  1.06$\pm$0.24  &  14.90$\pm$7.80  &  114.80$\pm$121.70  & 0.0007 & 0.0002 \\
					Pruning &  88.30$\pm$5.85  & 15    &  1.00$\pm$0.00  &  13.20$\pm$4.10  &  85.30$\pm$17.70  & 0.0005 & 0.0001 \\
					\midrule
					Without  &  88.10$\pm$6.88  & 13    &  1.06$\pm$0.24  &  8.30$\pm$7.20  &  63.90$\pm$84.10  & 0.0003 & 0.0002 \\
					Adaptive &  89.35$\pm$6.79  & 13    &  1.06$\pm$0.24  &  11.70$\pm$8.40  &  92.80$\pm$119.30  & 0.0007 & 0.0002 \\
					Memory &  88.57$\pm$7.13  & 15    &  1.06$\pm$0.24  &  10.40$\pm$8.40  &  82.90$\pm$113.50  & 0.0005 & 0.0001 \\
					\midrule
					Without  &  88.34$\pm$7.32  & 13    &  1.00$\pm$0.00  &  13.40$\pm$4.10  &  86.30$\pm$17.60  & 0.0003 & 0.0002 \\
					Soft Forgetting &  86.23$\pm$11.63  & 15    &  1.06$\pm$0.24  &  7.10$\pm$6.30  &  53.70$\pm$66.70  & 0.0007 & 0.0002 \\
					Mechanism &  88.87$\pm$5.93  & 16    &  1.06$\pm$0.24  &  12.70$\pm$6.10  &  93.60$\pm$79.40  & 0.0005 & 0.0001 \\
					\bottomrule
				\end{tabular}
			}
		\end{center}
		\label{tab:Tab1}%
	\end{table*}%
	
	\noindent\textbf{Related Works}: PNN \cite{progressiveneuralnetworks} suffers from complex structural burden because new nodes are added for each incoming task without any performance criteria while still relying on the conventional static depth trait. DEN \cite{DeepExpandable} makes use of problem-specific thresholds with opaque statistical interpretation and static network depth. \cite{Zhou_incrementallearning} is crafted under a shallow network structure and the structural learning mechanism is done with the absence of particular criterion. Furthermore, ODL \cite{OnlineDeepLearning} does not explore the issue of structural learning and simply works with a predefined network structure. 
	
	Three key differences exists between NADINE and ADL: 1) ADL is built upon a different-depth network structure via the weighted voting mechanism which limits its application only to classification problem; 2) the issue of catastrophic forgetting is handled in ADL via the localized learning concept to that of the winning layer determined from the layer weight adjusted using the penalty reward mechanism. The penalty and reward mechanism depends on the choice of step size and rather slow to react to the concept drift. NADINE, on the other hand, is equipped with the adaptive memory and soft forgetting concepts which assure relevance of network structure to already seen concept and no information loss in the incremental addition of new layers which has been a major challenge of MLP network's structural evolution; 3) ADL still adopts the standard SGD method whereas NADINE puts forward the soft forgetting approach governing the learning sensitivity of each layer during the SGD method according to the layer's relevance.   
	
	\begin{figure*}[h]
		\centering
		\includegraphics[scale=0.5]{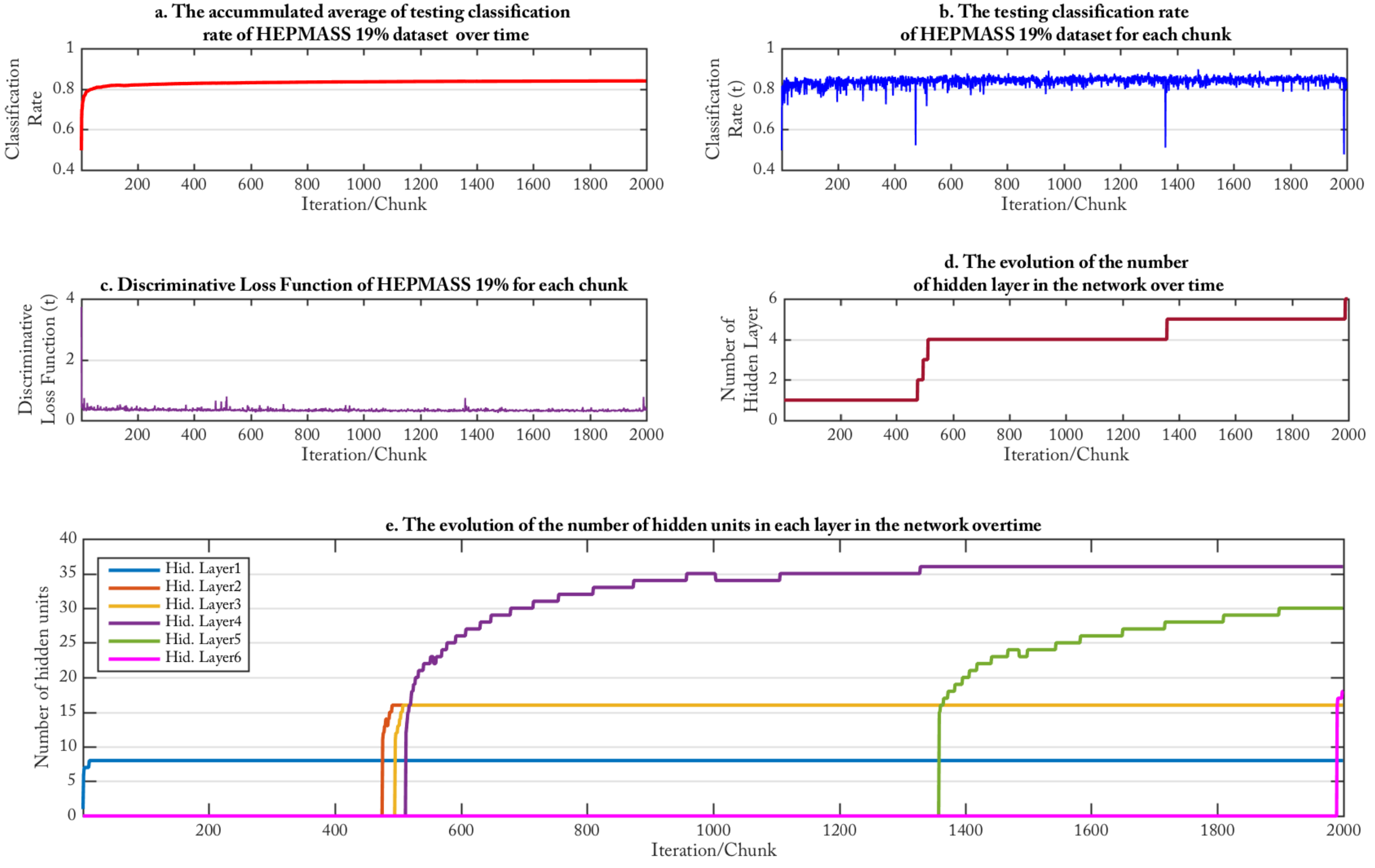}
		\caption{Performance metrics and the network evolution of NADINE in HEPMASS dataset problem}
		\label{fig:fig2}
	\end{figure*}
	
	\section{Conclusion}
	A novel self-organizing deep neural network, namely NADINE, is proposed in this paper. NADINE is built upon a flexible working principle where the structural learning process is automated with the absence of problem-specific user-defined parameters. It is framed in the standard multi-layer DNN structure where its structural learning strategy remains an open issue due to the catastrophic forgetting problem. The dynamic of network structure is governed by the estimation of bias and variance in the hidden unit level while the hidden layer is expandable using the drift detection method. The issue of catastrophic forgetting is resolved via the adaptive memory concept and the soft forgetting approach. Our numerical study in data stream classification and regression problems has confirmed the efficacy of NADINE in handling non-stationary data streams simulated in the prequential test-then-train protocol. It demonstrates performance improvement over its counterparts in all nine problems. 
	
	\section*{Acknowledgement}
	This research is partially funded by the National Research Foundation, Singapore under its AI Singapore programme [Award No.: AISG-RP-2018-004], its Industry Alignment Fund Pre-Positioning Programme in AME domain [Award No.: A19C1a0018] and the Data Science and Artificial Intelligence Research Center at Nanyang Technological University, Singapore. 
	\bibliographystyle{unsrt}  
	\bibliography{CIKM}
	
\end{document}